\documentclass{article}

\usepackage{graphicx}
\usepackage{subfigure}
\usepackage{makecell}
\usepackage{xcolor}
\usepackage{float}
\usepackage{booktabs}

\usepackage{hyperref}

\usepackage[accepted]{icml2020}

\begin{document}

\twocolumn[
\icmltitle{Chip Placement with Deep Reinforcement Learning}
\icmlsetsymbol{equal}{*}
\begin{icmlauthorlist}
\icmlauthor{Azalia Mirhoseini}{equal}
\icmlauthor{Anna Goldie}{equal}
\icmlauthor{Mustafa Yazgan}{}
\icmlauthor{Joe Jiang}{}
\icmlauthor{Ebrahim Songhori}{}
\icmlauthor{Shen Wang}{}
\icmlauthor{Young-Joon Lee}{}
\{azalia, agoldie, mustafay, wenjiej, esonghori, shenwang, youngjoonlee\}@google.com
\vspace{.1in}\\
\icmlauthor{Eric Johnson}{}
\icmlauthor{Omkar Pathak}{}
\icmlauthor{Sungmin Bae}{}
\icmlauthor{Azade Nazi}{}
\icmlauthor{Jiwoo Pak}{}
\icmlauthor{Andy Tong}{}
\icmlauthor{Kavya Srinivasa}{}
\icmlauthor{William Hang}{}
\icmlauthor{Emre Tuncer}{}
\icmlauthor{Anand Babu}{}
\icmlauthor{Quoc Le}{}
\icmlauthor{James Laudon}{}
\icmlauthor{Richard Ho}{}
\icmlauthor{Roger Carpenter}{}
\icmlauthor{Jeff Dean}{}
\end{icmlauthorlist}

\vskip 0.3in
]



\printAffiliationsAndNotice{\icmlEqualContribution} 
\begin{abstract}
In this work, we present a learning-based approach to chip placement, one of the most complex and time-consuming stages of the chip design process. Unlike prior methods, our approach has the ability to learn from past experience and improve over time. In particular, as we train over a greater number of chip blocks, our method becomes better at rapidly generating optimized placements for previously unseen chip blocks. To achieve these results, we pose placement as a Reinforcement Learning (RL) problem and train an agent to place the nodes of a chip netlist onto a chip canvas. To enable our RL policy to generalize to unseen blocks, we ground representation learning in the supervised task of predicting placement quality. By designing a neural architecture that can accurately predict reward across a wide variety of netlists and their placements, we are able to generate rich feature embeddings of the input netlists. We then use this architecture as the encoder of our policy and value networks to enable transfer learning. Our objective is to minimize PPA (power, performance, and area), and we show that, in under 6 hours, our method can generate placements that are superhuman or comparable on modern accelerator netlists, whereas existing baselines require human experts in the loop and take several weeks.


\end{abstract}

\section{Introduction}

Rapid progress in AI has been enabled by remarkable advances in computer systems and hardware, but with the end of Moore’s Law and Dennard scaling, the world is moving toward specialized hardware to meet AI’s exponentially growing demand for compute. However, today’s chips take years to design, leaving us with the speculative task of optimizing them for the machine learning (ML) models of 2-5 years from now. Dramatically shortening the chip design cycle would allow hardware to better adapt to the rapidly advancing field of AI. We believe that it is AI itself that will provide the means to shorten the chip design cycle, creating a symbiotic relationship between hardware and AI with each fueling advances in the other.

In this work, we present a learning-based approach to chip placement, one of the most complex and time-consuming stages of the chip design process. The objective is to place a netlist graph of macros (e.g., SRAMs) and standard cells (logic gates, such as NAND, NOR, and XOR) onto a chip canvas, such that power, performance, and area (PPA) are optimized, while adhering to constraints on placement density and routing congestion (described in Sections \ref{section:density} and \ref{section:congestion}). Despite decades of research on this problem, it is still necessary for human experts to iterate for weeks with the existing placement tools, in order to produce solutions that meet multi-faceted design criteria. The problem's complexity arises from the sizes of the netlist graphs (millions to billions of nodes), the granularity of the grids onto which these graphs must be placed, and the exorbitant cost of computing the true target metrics (many hours and sometimes over a day for industry-standard electronic design automation (EDA) tools to evaluate a single design). Even after breaking the problem into more manageable subproblems (e.g., grouping the nodes into a few thousand clusters and reducing the granularity of the grid), the state space is still orders of magnitude larger than recent problems on which learning-based methods have shown success.

To address this challenge, we pose chip placement as a Reinforcement Learning (RL) problem, where we train an agent (e.g., RL policy network) to optimize the placements. In each iteration of training, all of the macros of the chip block are sequentially placed by the RL agent, after which the standard cells are placed by a force-directed method \cite{forcedirected1972, dplace2008,mfar2005,kraftwerk2005,kraftwerk22008,fastplace2007,rql2007}. Training is guided by a fast-but-approximate reward signal for each of the agent's chip placements.

To our knowledge, the proposed method is the first placement approach with the ability to generalize, meaning that it can leverage what it has learned from placing previous netlists to generate placements for new unseen netlists. In particular, we show that, as our agent is exposed to a greater volume and variety of chips, it becomes both faster and better at generating optimized placements for new chip blocks, bringing us closer to a future in which chip designers are assisted by artificial agents with vast chip placement experience. 

 
 We believe that the ability of our approach to learn from experience and improve over time unlocks new possibilities for chip designers. We show that we can achieve superior PPA on real AI accelerator chips (Google TPUs), as compared to state-of-the-art baselines. Furthermore, our methods generate placements that are superior or comparable to human expert chip designers in under 6 hours, whereas the highest-performing alternatives require human experts in the loop and take several weeks for each of the dozens of blocks in a modern chip. Although we evaluate primarily on AI accelerator chips, our proposed method is broadly applicable to any chip placement optimization. 
 
\section{Related Work}
\label{section:related_work}

Global placement is a longstanding challenge in chip design, requiring multi-objective optimization over circuits of ever-growing complexity. Since the 1960s, many approaches have been proposed, so far falling into three broad categories: 1) partitioning-based methods, 2) stochastic/hill-climbing methods, and 3) analytic solvers. 

Starting in the 1960s, industry and academic labs took a partitioning-based approach to the global placement problem, proposing \cite{MinCutBreuer1977,TerminalPropagation1985,fiduccia1982}, as well as resistive-network based methods \cite{ResistiveNetwork1984,proud1988}. These methods are characterized by a divide-and-conquer approach; the netlist and the chip canvas are recursively partitioned until sufficiently small sub-problems emerge, at which point the sub-netlists are placed onto the sub-regions using optimal solvers. Such approaches are quite fast to execute and their hierarchical nature allows them to scale to arbitrarily large netlists. However, by optimizing each sub-problem in isolation, partitioning-based methods sacrifice quality of the global solution, especially routing congestion. Furthermore, a poor early partition may result in an unsalvageable end placement.

In the 1980s, analytic approaches emerged, but were quickly overtaken by stochastic / hill-climbing algorithms, particularly simulated annealing \cite{SimulatedAnnealing,DAC-1986-SechenS,dragon}. Simulated annealing (SA) is named for its analogy to metallurgy, in which metals are first heated and then gradually cooled to induce, or anneal, energy-optimal crystalline surfaces. SA applies random perturbations to a given placement (e.g., shifts, swaps, or rotations of macros), and then measures their effect on the objective function (e.g., half-perimeter wirelength described in Section~\ref{section:wirelength}). If the perturbation is an improvement, it is applied; if not, it is still applied with some probability, referred to as temperature. Temperature is initialized to a particular value and is then gradually annealed to a lower value. Although SA generates high-quality solutions, it is very slow and difficult to parallelize, thereby failing to scale to the increasingly large and complex circuits of the 1990s and beyond.

The 1990s-2000s were characterized by multi-level partitioning methods \cite{fengshui2005,capo2007}, as well as the resurgence of analytic techniques, such as force-directed methods \cite{dplace2008,mfar2005,kraftwerk2005,kraftwerk22008,fastplace2007,rql2007} and non-linear optimizers \cite{aplace22005,NTUPlacer06}. The renewed success of quadratic methods was due in part to algorithmic advances, but also to the large size of modern circuits (10-100 million nodes), which justified approximating the placement problem as that of placing nodes with zero area. However, despite the computational efficiency of quadratic methods, they are generally less reliable and produce lower quality solutions than their non-linear counterparts.

Non-linear optimization approximates cost using smooth mathematical functions, such as log-sum-exp \cite{logsumexp2001} and weighted-average \cite{weightedaverage2011} models for wirelength, as well as Gaussian \cite{ntuplace32008} and Helmholtz models for density. These functions are then combined into a single objective function using a Lagrange penalty or relaxation. Due to the higher complexity of these models, it is necessary to take a hierarchical approach, placing clusters rather than individual nodes, an approximation which degrades the quality of the placement.


The last decade has seen the rise of modern analytic techniques, including more advanced quadratic methods \cite{simpl2010,maple2012,complx2012,bonnplace2008,polar2013}, and more recently, electrostatics-based methods like ePlace \cite{EPlace15} and RePlAce \cite{RePlAce19}. Modeling netlist placement as an electrostatic system, ePlace \cite{EPlace15} proposed a new formulation of the density penalty where each node (macro or standard cell) of the netlist is analogous to a positively charged particle whose area corresponds to its electric charge. In this setting, nodes repel each other with a force proportional to their charge (area), and the density function and gradient correspond to the system's potential energy. Variations of this electrostatics-based approach have been proposed to address standard-cell placement \cite{EPlace15} and mixed-size placement \cite{EPlacemixed15,EPLace3D16}. RePlAce \cite{RePlAce19} is a recent state-of-the-art mixed-size placement technique that further optimizes ePlace's density function by introducing a local density function, which tailors the penalty factor for each individual bin size. Section~\ref{section:results} compares the performance of the state-of-the-art RePlAce algorithm against our approach. 

Recent work \cite{cnnplacer19} proposes training a model to predict the number of Design Rule Check (DRC) violations for a given macro placement. DRCs are rules that ensure that the placed and routed netlist adheres to tape-out requirements. To generate macro placements with fewer DRCs, \cite{cnnplacer19} use the predictions from this trained model as the evaluation function in simulated annealing. While this work represents an interesting direction, it reports results on netlists with no more than 6 macros, far fewer than any modern block, and the approach does not include any optimization during the place and the route steps. Due to the optimization, the placement and the routing can change dramatically, and the actual DRC will change accordingly, invalidating the model prediction. In addition, although adhering to the DRC criteria is a necessary condition, the primary objective of macro placement is to optimize for wirelength, timing (e.g. Worst Negative Slack (WNS) and Total Negative Slack (TNS)), power, and area, and this work does not even consider these metrics. 

To address this classic problem, we propose a new category of approach: end-to-end learning-based methods. This type of approach is most closely related to analytic solvers, particularly non-linear ones, in that all of these methods optimize an objective function via gradient updates. However, our approach differs from prior approaches in its ability to learn from past experience to generate higher-quality placements on new chips. Unlike existing methods that optimize the placement for each new chip from scratch, our work leverages knowledge gained from placing prior chips to become better over time. In addition, our method enables direct optimization of the target metrics, such as wirelength, density, and congestion, without having to define convex approximations of those functions as is done in other approaches \cite{RePlAce19,EPlace15}. Not only does our formulation make it easy to incorporate new cost functions as they become available, but it also allows us to weight their relative importance according to the needs of a given chip block (e.g., timing-critical or power-constrained).

Domain adaptation is the problem of training policies that can learn across multiple experiences and transfer the acquired knowledge to perform better on new unseen examples. In the context of chip placement, domain adaptation involves training a policy across a set of chip netlists and applying that trained policy to a new unseen netlist. Recently, domain adaptation for combinatorial optimization has emerged as a trend \cite{zhou2019gdp,REGAL19,Placeto18}. While the focus in prior work has been on using domain knowledge learned from previous examples of an optimization problem to speed up policy training on new problems, we propose an approach that, for the first time, enables the generation of higher quality results by leveraging past experience. Not only does our novel domain adaptation produce better results, it also reduces the training time 8-fold compared to training the policy from scratch.

\section{Methods}
\label{section:methods}

\subsection{Problem Statement}
\label{section:problem_statement}

In this work, we target the chip placement optimization problem, in which the objective is to map the nodes of a netlist (the graph describing the chip) onto a chip canvas (a bounded 2D space), such that final power, performance, and area (PPA) is optimized. In this section, we describe an overview of how we formulate the problem as a reinforcement learning (RL) problem, followed by a detailed description of the reward function, action and state representations, policy architecture, and policy updates. 


\subsection{Overview of Our Approach}
\label{section:overview}
\begin{figure*}
    \centering
    \includegraphics[width=0.98\textwidth]{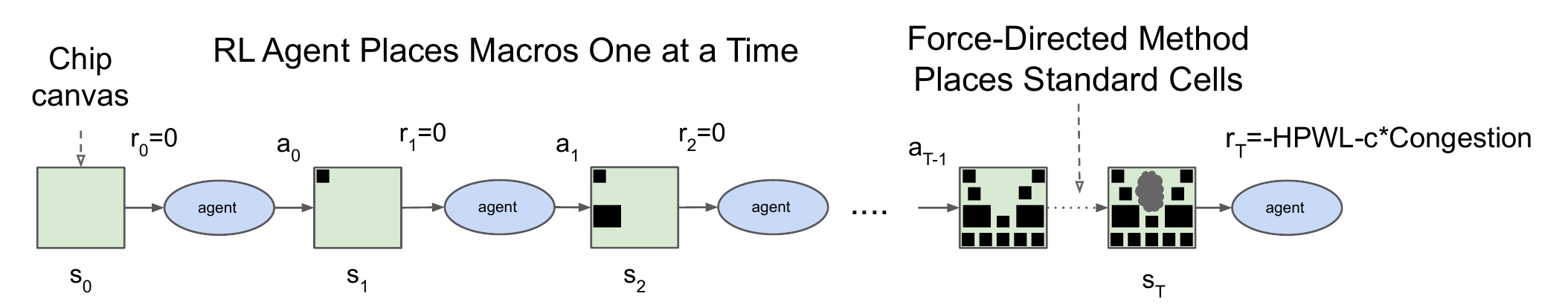}
    \caption{The RL agent (i.e., the policy network) places macros one at a time. Once all macros are placed, the standard cells are placed using a force-directed method. The reward, a linear combination of the approximate wirelength and congestion, are calculated and passed to the agent to optimize its parameters for the next iteration.}
    \label{fig:placement-actions}
\end{figure*}
We take a deep reinforcement learning approach to the placement problem, where an RL agent (policy network) sequentially places the macros; once all macros are placed, a force-directed method is used to produce a rough placement of the standard cells, as shown in Figure~\ref{fig:placement-actions}. RL problems can be formulated as Markov Decision Processes (MDPs), consisting of four key elements:
\begin{itemize}
  \item states: the set of possible states of the world (e.g., in our case, every possible partial placement of the netlist onto the chip canvas).
  \item actions: the set of actions that can be taken by the agent (e.g., given the current macro to place, the available actions are the set of all the locations in the discrete canvas space (grid cells) onto which that macro can be placed without violating any hard constraints on density or blockages).
  \item state transition: given a state and an action, this is the probability distribution over next states.
  \item reward: the reward for taking an action in a state. (e.g., in our case, the reward is 0 for all actions except the last action where the reward is a negative weighted sum of proxy wirelength and congestion, subject to density constraints as described in Section~\ref{section:reward}).
\end{itemize}

In our setting, at the initial state, $s_0$, we have an empty chip canvas and an unplaced netlist. The final state $s_T$ corresponds to a completely placed netlist. At each step, one macro is placed. Thus, $T$ is equal to the total number of macros in the netlist. At each time step $t$, the agent begins in state ($s_t$), takes an action ($a_t$), arrives at a new state ($s_{t+1}$), and receives a reward ($r_t$) from the environment (0 for $t<T$ and negative proxy cost for $t=T$). 

We define $s_t$ to be a concatenation of features representing the state at time $t$, including a graph embedding of the netlist (including both placed and unplaced nodes), a node embedding of the current macro to place, metadata about the netlist (Section~\ref{section:domain_transfer}), and a mask representing the feasibility of placing the current node onto each cell of the grid (Section~\ref{section:density}).

The action space is all valid placements of the $t^{th}$ macro, which is a function of the density mask described in section~\ref{section:density}. Action $a_t$ is the cell placement of the $t^{th}$ macro that was chosen by the RL policy network. 

$s_{t+1}$ is the next state, which includes an updated representation containing information about the newly placed macro, an updated density mask, and an embedding for the next node to be placed. 

In our formulation, $r_t$ is $0$ for every time step except for the final $r_T$, where it is a weighted sum of approximate wirelength and congestion as described in Section~\ref{section:reward}.

Through repeated episodes (sequences of states, actions, and rewards), the policy network learns to take actions that will maximize cumulative reward. We use Proximal Policy Optimization (PPO)~\cite{ppo17} to update the parameters of the policy network, given the cumulative reward for each placement. 

In this section, we define the reward $r$, state $s$, actions $a$, policy network architecture $\pi_{\theta}(a|s)$ parameterized by $\theta$, and finally the optimization method we use to train those parameters.

\subsection{Reward}\label{section:reward}
Our goal in this work is to minimize power, performance and area, subject to constraints on routing congestion and density. Our true reward is the output of a commercial EDA tool, including wirelength, routing congestion, density, power, timing, and area. However, RL policies require 100,000s of examples to learn effectively, so it is critical that the reward function be fast to evaluate, ideally running in a few milliseconds. In order to be effective, these approximate reward functions must also be positively correlated with the true reward. Therefore, a component of our cost is wirelength, because it is not only much cheaper to evaluate, but also correlates with power and performance (timing). We define approximate cost functions for both wirelength and congestion, as described in Section~\ref{section:wirelength} and Section~\ref{section:congestion}, respectively.

To combine multiple objectives into a single reward function, we take the weighted sum of proxy wirelength and congestion where the weight can be used to explore the trade-off between the two metrics.

While we treat congestion as a soft constraint (i.e., lower congestion improves the reward function), we treat density as a hard constraint, masking out actions (grid cells to place nodes onto) whose density exceeds the target density, as described further in section~\ref{section:density}.

To keep the runtime per iteration small, we apply several approximations to the calculation of the reward function: 

\begin{enumerate}
  \item We group millions of standard cells into a few thousand clusters using hMETIS~\cite{hmetis1998}, a partitioning technique based on the normalized minimum cut objective. Once all the macros are placed, we use force-directed methods to place the standard cell clusters, as described in section~\ref{section:standard_cell_placement}. Doing so enables us to achieve an approximate but fast standard cell placement that facilitates policy network optimization.
  \item We discretize the grid to a few thousand grid cells and place the center of macros and standard cell clusters onto the center of the grid cells.
  \item When calculating wirelength, we make the simplifying assumption that all wires leaving a standard cell cluster  originate at the center of the cluster.
  \item To calculate routing congestion cost, we only consider the average congestion of the top 10\% most congested grid cells, as described in Section~\ref{section:congestion}.
\end{enumerate}

\subsubsection{Wirelength}
\label{section:wirelength}
Following the literature~\cite{hpwl1991}, we employ half-perimeter wirelength (HPWL), the most commonly used approximation for wirelength. HPWL is defined as the half-perimeter of the bounding boxes for all nodes in the netlist. The HPWL for a given net (edge) $i$ is shown in the equation below:

\begin{eqnarray}
  \label{eqn:hpwl}
  HPWL(i) = & (MAX_{b \in i}\{x_b\} - MIN_{b \in i}\{x_b\} + 1) \\ \nonumber 
         + & (MAX_{b \in i}\{y_b\} - MIN_{b \in i}\{y_b\} + 1) 
\end{eqnarray}

Here $x_b$ and $y_b$ show the x and y coordinates of the end points of net $i$. The overall HPWL cost is then calculated by taking the normalized sum of all half-perimeter bounding boxes, as shown in Equation \ref{eqn:wirelength_cost}. $q(i)$ is a normalization factor which improves the accuracy of the estimate by increasing the wirelength cost as the number of nodes increases, where $N_{netlist}$ is the number of nets. 

\begin{eqnarray}
  \label{eqn:wirelength_cost}
  HPWL(netlist) = \sum_{i=1}^{N_{netlist}}q(i)*HPWL(i)
\end{eqnarray}

Intuitively, the HPWL for a given placement is roughly the length of its Steiner tree~\cite{gilbert1968steiner}, which is a lower bound on routing cost.

Wirelength also has the advantage of correlating with other important metrics, such as power and timing. Although we don't optimize directly for these other metrics, we observe high performance in power and timing (as shown in Table~\ref{table:replace-comparison}).

\subsubsection{Selection of grid rows and columns}
\label{section:selecting_rows_and_columns}
Given the dimensions of the chip canvas, there are many choices to discretize the 2D canvas into grid cells. This decision impacts the difficulty of optimization and the quality of the final placement. We limit the maximum number of rows and columns to 128. We treat choosing the optimal number of rows and columns as a bin-packing problem and rank different combinations of rows and columns by the amount of wasted space they incur. We use an average of 30 rows and columns in the experiments described in Section~\ref{section:results}.



\subsubsection{Selection of macro order}
\label{section:selecting_macro_order}
To select the order in which the macros are placed, we sort macros by descending size and break ties using a topological sort. By placing larger macros first, we reduce the chance of there being no feasible placement for a later macro. The topological sort can help the policy network learn to place connected nodes close to one another. Another potential approach would be to learn to jointly optimize the ordering of macros and their placement, making the choice of which node to place next part of the action space. However, this enlarged action space would significantly increase the complexity of the problem, and we found that this heuristic worked in practice.

\subsubsection{Standard cell placement}
\label{section:standard_cell_placement}
To place standard cell clusters, we use an approach similar to classic force-directed methods~\cite{hpwl1991}. We represent the netlist as a system of springs that apply force to each node, according to the $weight\times distance$ formula, causing tightly connected nodes to be attracted to one another. We also introduce a repulsive force between overlapping nodes to reduce placement density. After applying all forces, we move nodes in the direction of the force vector. To reduce oscillations, we set a maximum distance for each move.

\subsubsection{Routing congestion}
\label{section:congestion}
We also followed convention in calculating proxy congestion~\cite{MAPLE12}, using a simple deterministic routing based on the locations of the driver and loads on the net. The routed net occupies a certain amount of available routing resources (determined by the underlying semiconductor fabrication technology) for each grid cell which it passes through. We keep track of vertical and horizontal allocations in each grid cell separately. To smoothe the congestion estimate, we run $5\times 1$ convolutional filters in both the vertical and horizontal direction. After all nets are routed, we take the average of the top $10\%$ congestion values, drawing inspiration from the ABA10 metric in MAPLE~\cite{MAPLE12}. The congestion cost in Equation~\ref{eqn:reward} is the top $10\%$ average congestion calculated by this process.

\subsubsection{Density}
\label{section:density}
We treat density as a hard constraint, disallowing the policy network from placing macros in locations which would cause density to exceed the target ($max_{density}$) or which would result in infeasible macro overlap. This approach has two benefits: (1) it reduces the number of invalid placements generated by the policy network, and (2) it reduces the search space of the optimization problem, making it more computationally tractable.

A feasible standard cell cluster placement should meet the following criterion: the density of placed items in each grid cell should not exceed a given target density threshold ($max_{density}$). We set this threshold to be $0.6$ in our experiments. To meet this constraint, during each RL step, we calculate the current density mask, a binary $m\times n$ matrix that represents grid cells onto which we can place the center of the current node without violating the density threshold criteria. Before choosing an action from the policy network output, we first take the dot product of the mask and the policy network output and then take the argmax over feasible locations. This approach prevents the policy network from generating placements with overlapping macros or dense standard cell areas.

We also enable blockage-aware placements (such as clock straps) by setting the density function of the blocked areas to $1$.

\subsubsection{Postprocessing}
\label{section:postprocessing}
To prepare the placements for evaluation by commercial EDA tools, we perform a greedy legalization step to snap macros onto the nearest legal position while honoring the minimum spacing constraints. We then fix the macro placements and use an EDA tool to place the standard cells and evaluate the placement.  


\subsection{Action Representation}
\label{section:actions}
For policy optimization purposes, we convert the canvas into a $m\times n$ grid. Thus, for any given state, the action space (or the output of the policy network) is the probability distribution of placements of the current macro over the $m\times n$ grid. The action is the argmax of this probability distribution.

\subsection{State Representation}
\label{section:states}
Our state contains information about the netlist graph (adjacency matrix), its node features (width, height, type, etc.), edge features (number of connections), current node (macro) to be placed, and metadata of the netlist and the underlying technology (e.g., routing allocations, total number of wires, macros, and standard cell clusters, etc.). In the following section, we discuss how we process these features to learn effective representations for the chip placement problem.

\section{Domain Transfer: Learning Better Chip Placements from Experience}
\label{section:domain_transfer}

\begin{figure*}
    \centering
    \includegraphics[width=0.95\textwidth]{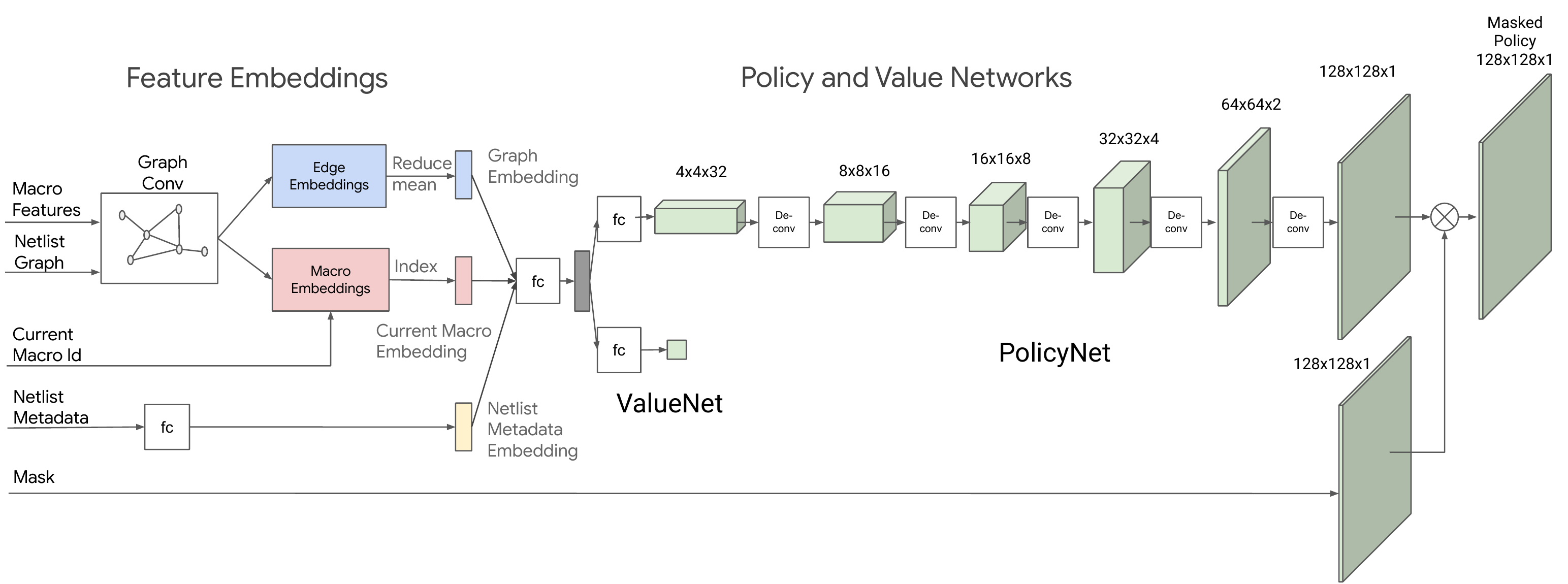}
    \caption{Policy and value network architecture. An embedding layer encodes information about the netlist adjacency, node features, and the current macro to be placed. The policy and value networks then output a probability distribution over available placement locations and an estimate of the expected reward for the current placement, respectively.}
    \label{fig:policy-architecture}
\end{figure*}

Our goal is to develop RL agents that can generate higher quality results as they gain experience placing chips. We can formally define the placement objective function as follows:

\begin{eqnarray}
  \label{eqn:objective}
  J(\theta, G) = \frac{1}{K}\sum_{g\sim G}{E_{g,p\sim\pi_\theta}}[R_{p,g}]
\end{eqnarray}

 Here $J(\theta, G)$ is the cost function.  The agent is parameterized by $\theta$. The dataset of netlist graphs of size $K$ is denoted by $G$ with each individual netlist in the dataset written as $g$. $R_{p,g}$ is the episode reward of a placement $p$ drawn from the policy network applied to netlist $g$.

\begin{eqnarray}\label{eqn:reward}
R_{p,g} = -Wirelength(p,g) - \lambda~Congestion(p,g)  \\ \nonumber
S.t. ~density(p,g) \le max_{density}
\end{eqnarray}

Equation \ref{eqn:reward} shows the reward that we used for policy network optimization, which is the negative weighted average of wirelength and congestion, subject to density constraints. The reward is explained in detail in Section~\ref{section:reward}. In our experiments, congestion weight $\lambda$ is set to 0.01 and the max density threshold is set to 0.6.

\subsection{A Supervised Approach to Enable Transfer Learning}

We propose a novel neural architecture that enables us to train domain-adaptive policies for chip placement. Training such a policy network is a challenging task since the state space encompassing all possible placements of all possible chips is immense. Furthermore, different netlists and grid sizes can have very different properties, including differing numbers of nodes, macro sizes, graph topologies, and canvas widths and heights. To address this challenge, we first focused on learning rich representations of the state space. Our intuition was that a policy network architecture capable of transferring placement optimization across chips should also be able to encode the state associated with a new unseen chip into a meaningful signal at inference time. We therefore proposed training a neural network architecture capable of predicting reward on new netlists, with the ultimate goal of using this architecture as the encoder layer of our policy network.

To train this supervised model, we needed a large dataset of chip placements and their corresponding reward labels. We therefore created a dataset of 10,000 chip placements where the input is the state associated with a given placement and the label is the reward for that placement (wirelength and congestion). We built this dataset by first picking 5 different accelerator netlists and then generating 2,000 placements for each netlist. To create diverse placements for each netlist, we trained a vanilla policy network at various congestion weights (ranging from 0 to 1) and random seeds, and collected snapshots of each placement during the course of policy training. An untrained policy network starts off with random weights and the generated placements are of low quality, but as the policy network trains, the quality of generated placements improves, allowing us to collect a diverse dataset with placements of varying quality. 

To train a supervised model that can accurately predict wirelength and congestion labels and generalize to unseen data, we developed a novel graph neural network architecture that embeds information about the netlist. The role of graph neural networks is to distill information about the type and connectivity of a node within a large graph into low-dimensional vector representations which can be used in downstream tasks. Some examples of such downstream tasks are node classification~\cite{nazi2019gap}, device placement~\cite{zhou2019gdp}, link prediction~\cite{zhang2018link}, and Design Rule Violations (DRCs) prediction~\cite{RouteNet18}.

 We create a vector representation of each node by concatenating the node features. The node features include node type, width, height, and x and y coordinates. We also pass node adjacency information as input to our algorithm. We then repeatedly perform the following updates: 1) each edge updates its representation by applying a fully connected network to an aggregated representation of intermediate node embeddings, and 2) each node updates its representation by taking the mean of adjacent edge embeddings. The node and edge updates are shown in Equation \ref{eq:graphalg}.
 
\begin{eqnarray}\label{eq:graphalg}
 e_{ij} = & fc_1(concat(fc_0(v_i) | fc_0(v_j) | w^e_{ij})) \\ \nonumber
 v_i = & mean_{j\in \mathcal{N}(v_i)}(e_{ij})
\end{eqnarray}

Node embeddings are denoted by $v_i$s for $1<=i<=N$, where $N$ is the total number of macros and standard cell clusters. Vectorized edges connecting nodes $v_i$ and $v_j$ are represented as $e_{ij}$. Both edge ($e_{ij}$) and node ($v_i$) embeddings are randomly initialized and are 32-dimensional. $fc_0$ is a $32\times32$, $fc_1$ is a $65\times 32$ feedforward network and $w^e_{ij}$s are learnable $1x1$ weights corresponding to edges. $\mathcal{N}(v_i)$ shows the neighbors of $v_i$. The outputs of the algorithm are the node and edge embeddings.



Our supervised model consists of: (1) The graph neural network described above that embeds information about node types and the netlist adjacency matrix. (2) A fully connected feedforward network that embeds the metadata, including information about the underlying semiconductor technology (horizontal and vertical routing capacity), the total number of nets (edges), macros, and standard cell clusters, canvas size and number of rows and columns in the grid. (3) A fully connected feedforward network (the prediction layer) whose input is a concatenation of the netlist graph and metadata embedding and whose output is the reward prediction. The netlist graph embedding is created by applying a reduce mean function on the edge embeddings. The supervised model is trained via regression to minimize the weighted sum of the mean squared loss of wirelength and congestion. 

This supervised task allowed us to find the features and architecture necessary to generalize reward prediction across netlists. To incorporate this architecture into our policy network, we removed the prediction layer and then used it as the encoder component of the policy network as shown in Figure \ref{fig:policy-architecture}.

To handle different grid sizes corresponding to different choices of rows and columns, we set the grid size to $128\times 128$, and mask the unused L-shaped section for grid sizes smaller than 128 rows and columns.

To place a new test netlist at inference time, we load the pre-trained weights of the policy network and apply it to the new netlist. We refer to placements generated by a pre-trained policy network with no finetuning as zero-shot placements. Such a placement can be generated in less than a second, because it only requires a single inference step of the pre-trained policy network. We can further optimize placement quality by finetuning the policy network. Doing so gives us the flexibility to either use the pre-trained weights (that have learned a rich representation of the input state) or further finetune these weights to optimize for the properties of a particular chip netlist.  


\begin{figure*}[t]
    \centering
    \includegraphics[width=0.7\textwidth]{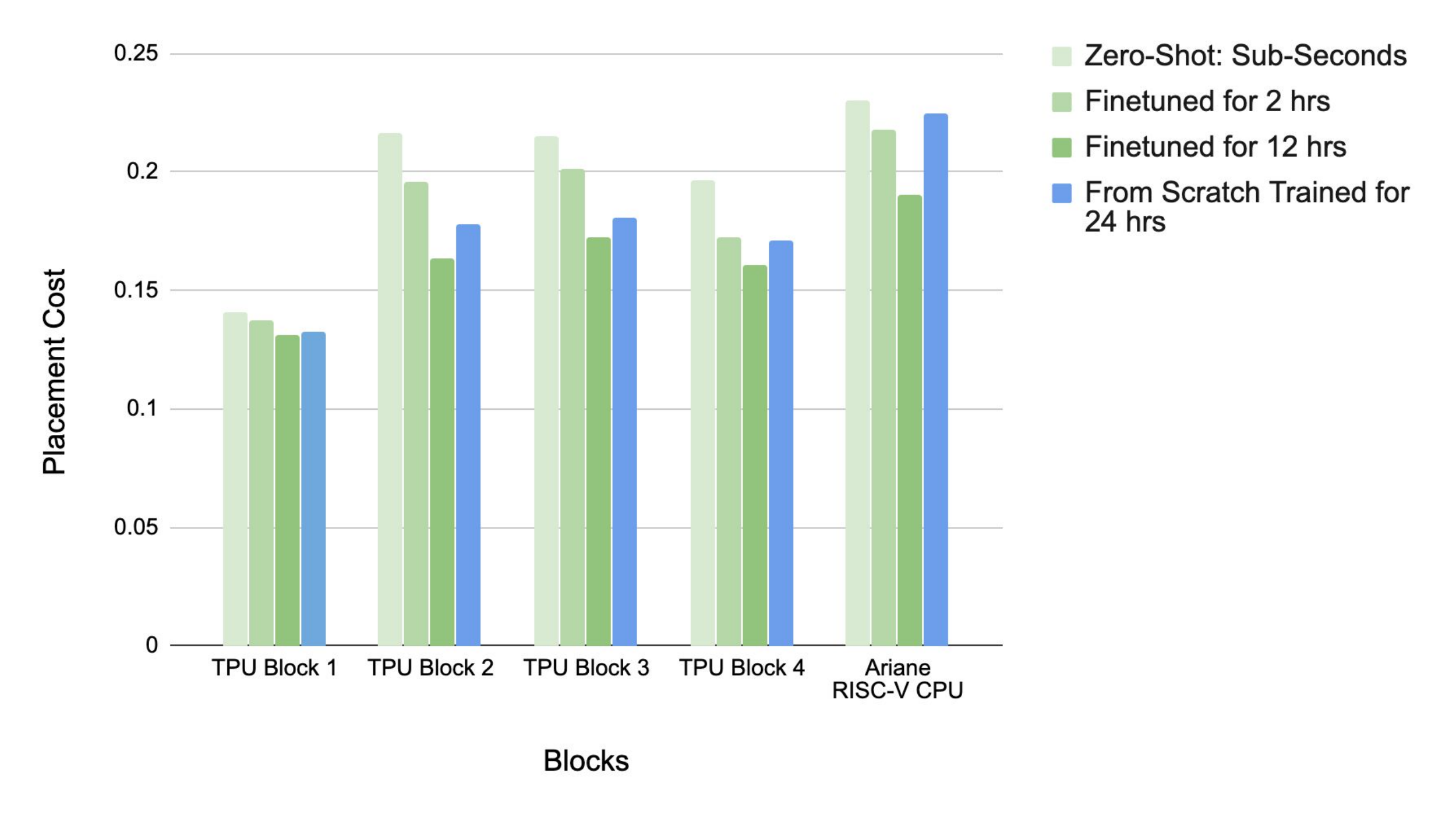}
    \caption{Domain adaptation results. For each block, the zero-shot results, as well as the finetuned results after 2 and 6 hours of training are shown. We also include results for policies trained from scratch. As can be seen in the table, the pre-trained policy network consistently outperforms the policy network that was trained from scratch, demonstrating the effectiveness of learning from training data offline.}
    \label{fig:generalizationresults}
\end{figure*}
\begin{figure*}[t]
    \centering
    \includegraphics[width=0.6\textwidth]{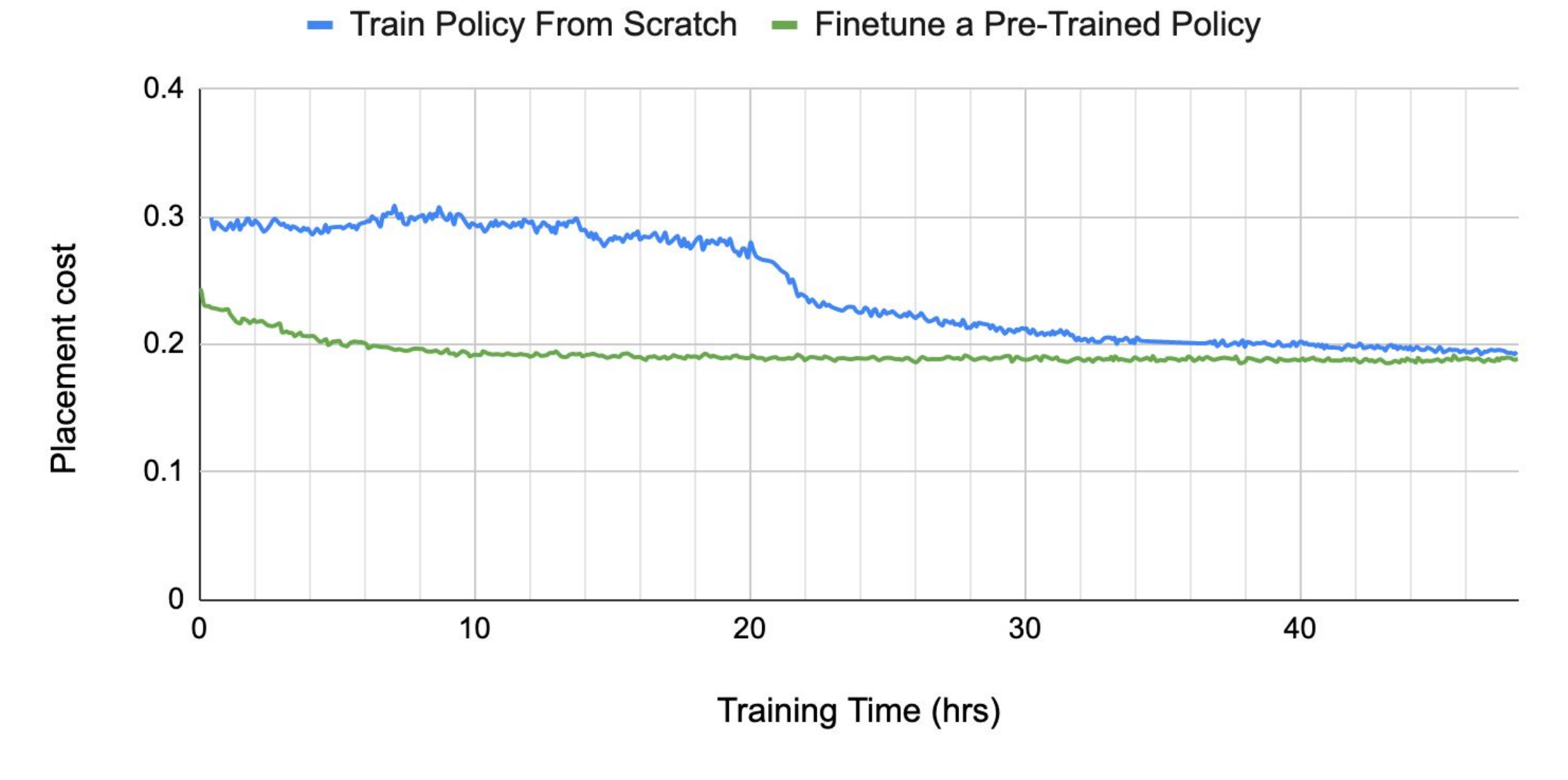}
    \caption{Convergence plots for training a policy network from scratch vs. finetuning a pre-trained policy network for a block of Ariane.}
    \label{fig:convergenceresults}
\end{figure*}

\subsection{Policy Network Architecture}
Figure \ref{fig:policy-architecture} depicts an overview of the policy network (modeled by $\pi_{\theta}$ in Equation \ref{eqn:objective}) and the value network architecture that we developed for chip placement. The inputs to these networks are the netlist graph (graph adjacency matrix and node features), the id of the current node to be placed, and the metadata of the netlist and the semiconductor technology. The netlist graph is passed through our proposed graph neural network architecture as described earlier. This graph neural network generates embeddings of (1) the partially placed graph and (2) the current node. We use a simple feedforward network to embed (3) the metadata. These three embedding vectors are then concatenated to form the state embedding, which is passed to a feedforward neural network. The output of the feedforward network is then fed into the policy network (composed of 5 deconvolutions~\footnote{The deconvolutions layers have a 3x3 kernel size with stride 2 and 16, 8, 4, 2, and 1 filter channels respectively.} and Batch Normalization layers) to generate a probability distribution over actions and passed to a value network (composed of a feedforward network) to predict the value of the input state.  


\subsection{Policy Network Update: Training Parameters $\theta$}
\label{section:policy_update}
In Equation \ref{eqn:objective}, the objective is to train a policy network $\pi_{\theta}$ that maximizes the expected value ($E$) of the reward ($R_{p,g}$) over the policy network's placement distribution. To optimize the parameters of the policy network, we use Proximal Policy Optimization (PPO) \cite{ppo17} with a clipped objective as shown below: 
\begin{eqnarray}
  \label{eqn:cost}
  L^{CLIP}(\theta) = {\hat{E_t}}[min(r_t(\theta)\hat{A_t}, clip(r_t(\theta), 1 - \epsilon, 1 + \epsilon)\hat{A_t})]\nonumber
\end{eqnarray}
\vspace{-0.05in}
where $\hat{E_t}$ represents the expected value at timestep $t$, $r_t$ is the ratio of the new policy and the old policy, and $\hat{A_t}$ is the estimated advantage at timestep $t$.

\section{Results}\label{section:results}
In this section, we evaluate our method and answer the following questions: Does our method enable domain transfer and learning from experience? What is the impact of using pre-trained policies on the quality of result? How does the quality of the generated placements compare to state-of-the-art baselines? We also inspect the visual appearance of the generated placements and provide some insights into why our policy network made those decisions.

\begin{figure*}[t]
    \centering
    \begin{subfigure}
        \centering
        \includegraphics[width=0.45\linewidth]{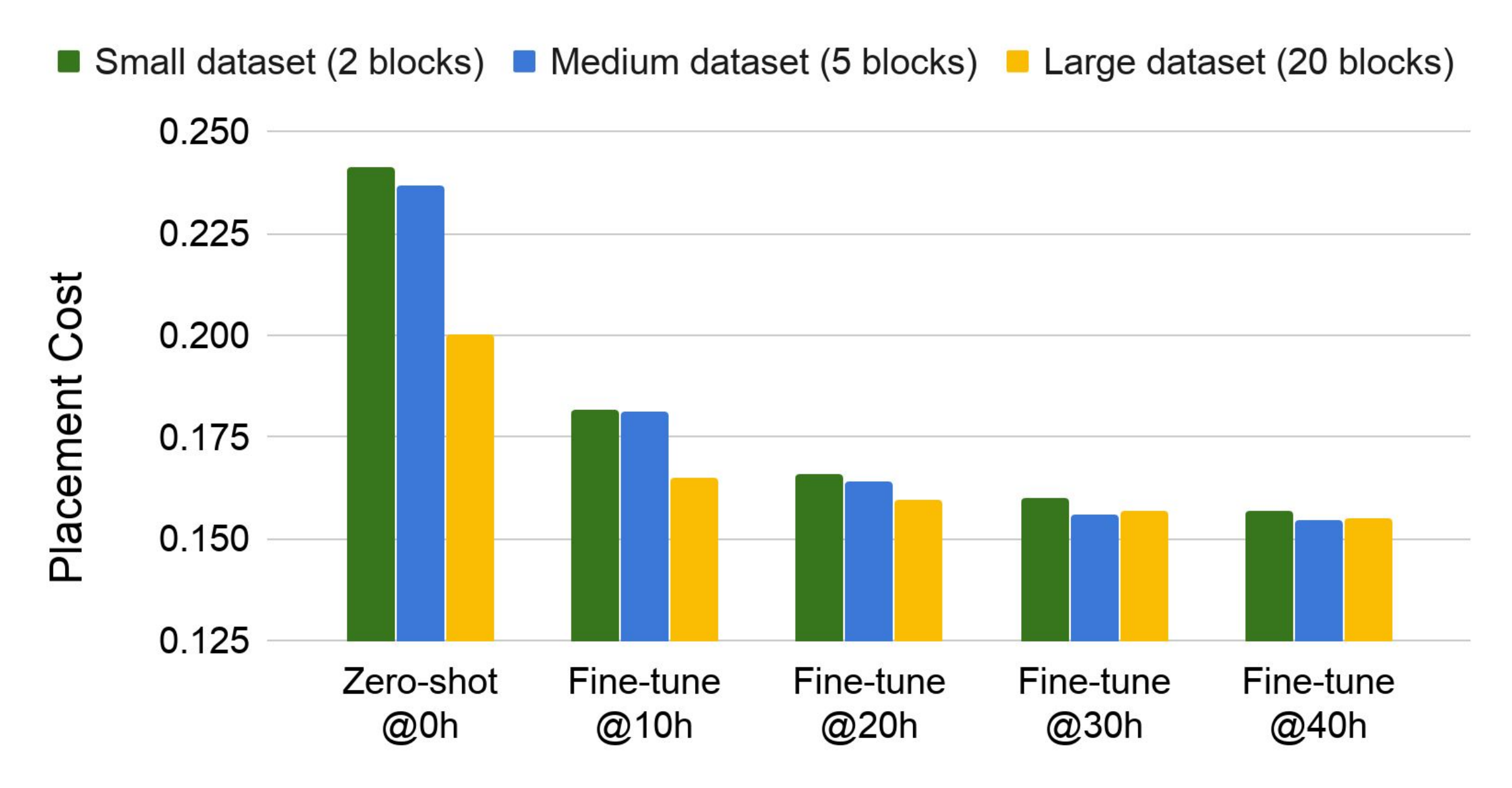}
    \end{subfigure}
    \begin{subfigure}
        \centering  
        \includegraphics[width=0.45\linewidth]{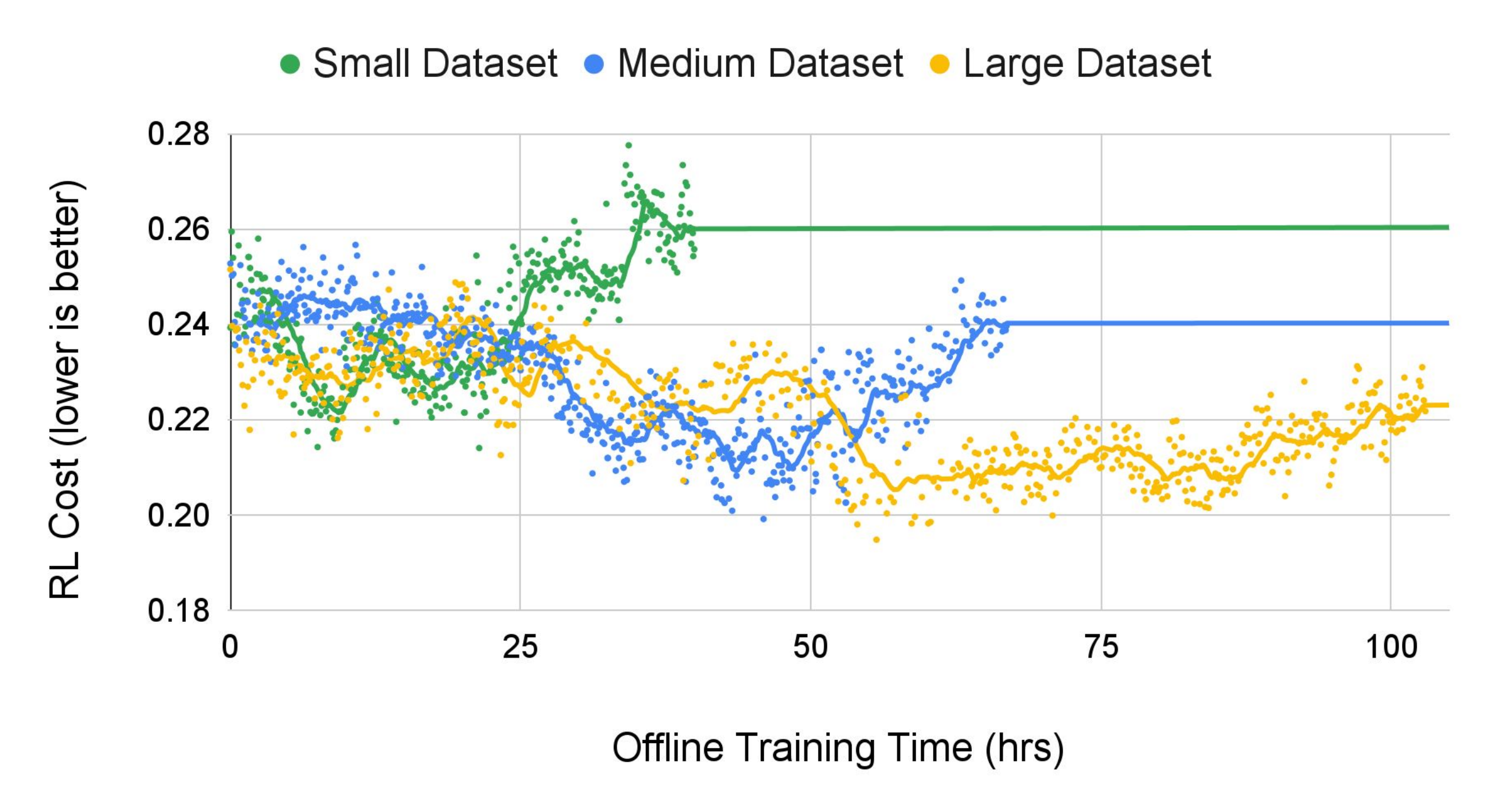}
    \end{subfigure}
    \caption{We pre-train the policy network on three different training datasets (the small dataset is a subset of the medium one, and the medium dataset is a subset of the large one). We then finetune this pre-trained policy network on the same test block and report cost at various training durations (shown on the left of the figure). As the dataset size increases, both the quality of generated placements and time to convergence on the test block improve. The right figure shows  evaluation curves for policies trained on each dataset (each dot in the right figure shows the cost of the placement generated by the policy under training).}
    \label{fig:trainingsize}
\end{figure*}
\begin{figure*}[ht]
     \centering
     \begin{subfigure}
        \centering
        \includegraphics[width=0.35\textwidth]{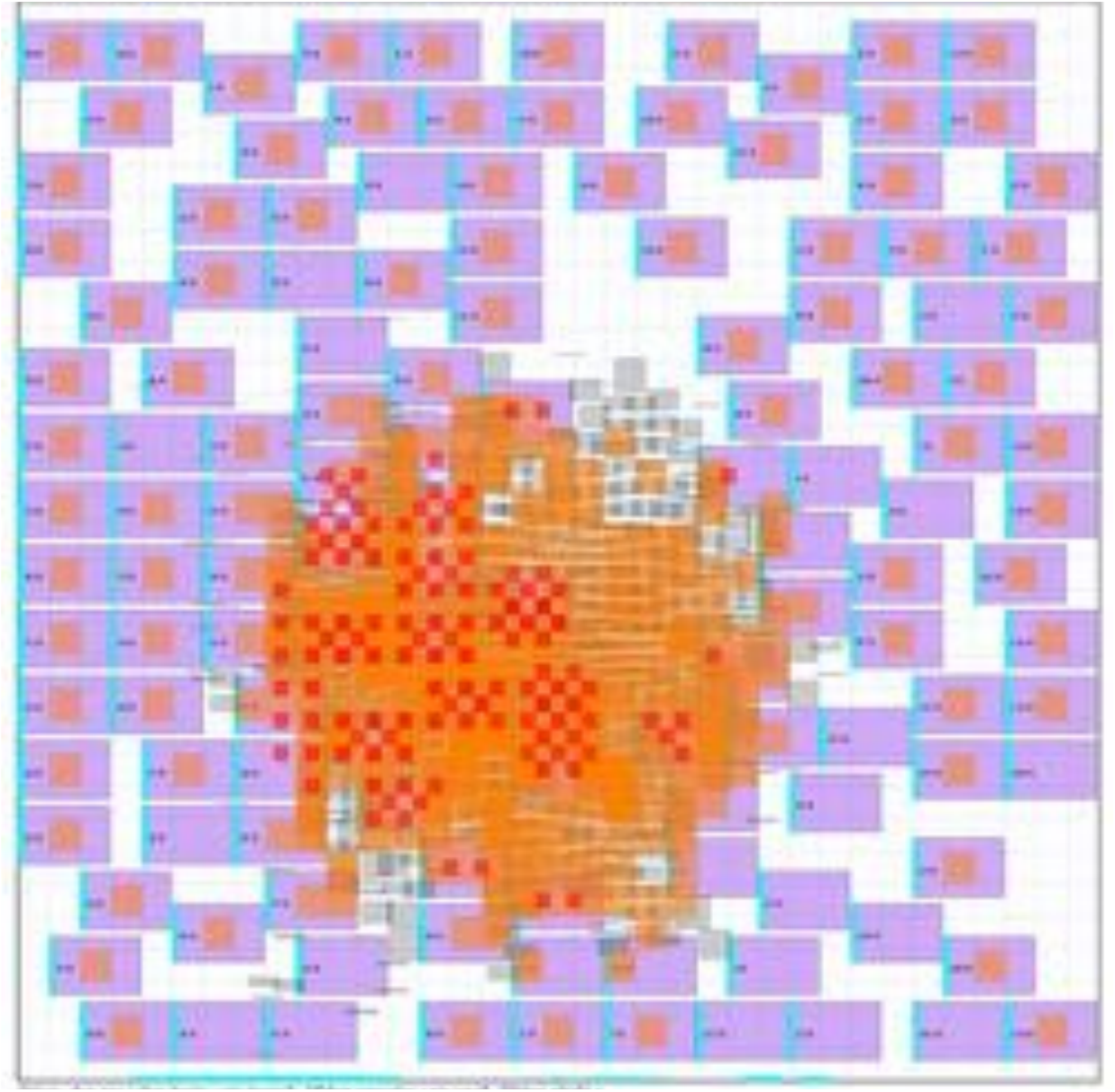}
     \end{subfigure}
     \begin{subfigure}
        \centering
        \includegraphics[width=0.35\textwidth]{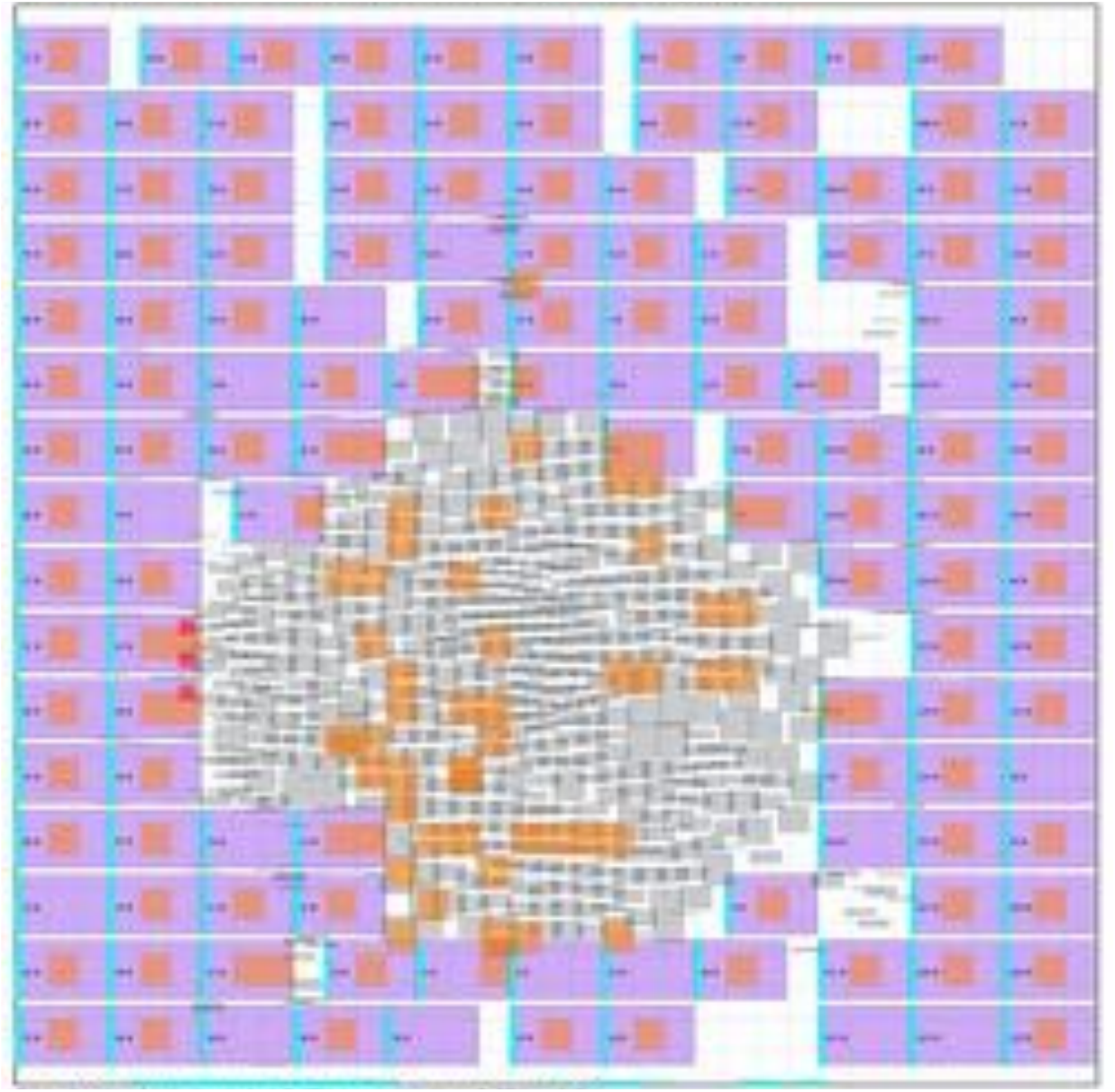}
     \end{subfigure}
        \caption{Visualization of placements. On the left, zero-shot placements from the pre-trained policy and on the right, placements from the finetuned policy are shown. The zero-shot policy placements are generated at inference time on a previously unseen chip. The pre-trained policy network (with no fine-tuning) places the standard cells in the center of the canvas surrounded by macros, which is already quite close to the optimal arrangement and in line with the intuitions of physical design experts.}
        \label{fig:zeroshot-vs-finetuned}
\end{figure*}

\subsection{Transfer Learning Results}

Figure \ref{fig:generalizationresults} compares the quality of placements generated using pre-trained policies to those generated by training the policy network from scratch. Zero-shot means that we applied a pre-trained policy network to a new netlist with no finetuning, yielding a placement in less than one second. We also show results where we finetune the pre-trained policy network on the details of a particular design for 2 and 12 hours. The policy network trained from scratch takes much longer to converge, and even after 24 hours, the results are worse than what the finetuned policy network achieves after 12 hours, demonstrating that the learned weights and exposure to many different designs are helping us to achieve higher quality placements for new designs in less time.  

Figure \ref{fig:convergenceresults} shows the convergence plots for training from scratch vs. training from a pre-trained policy network for Ariane RISC-V CPU. The pre-trained policy network starts with a lower placement cost at the beginning of the finetuning process. Furthermore, the pre-trained policy network converges to a lower placement cost and does so more than 30 hours faster than the policy network that was trained from scratch.


\begin{figure*}[t]
    \centering
    \begin{subfigure}
        \centering
        \includegraphics[width=0.45\textwidth]{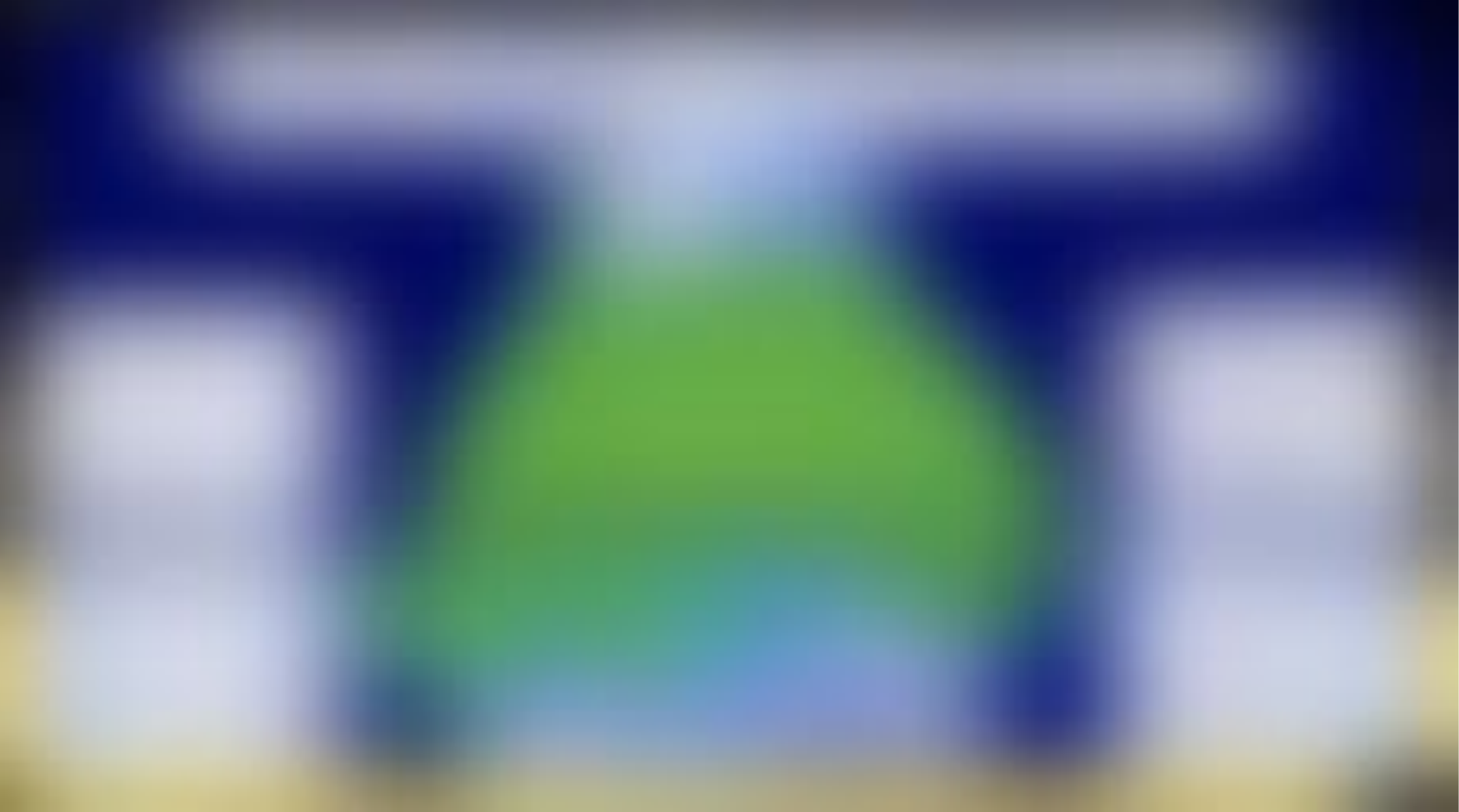}
    \end{subfigure}
    \begin{subfigure}
        \centering
         \includegraphics[width=0.45\textwidth]{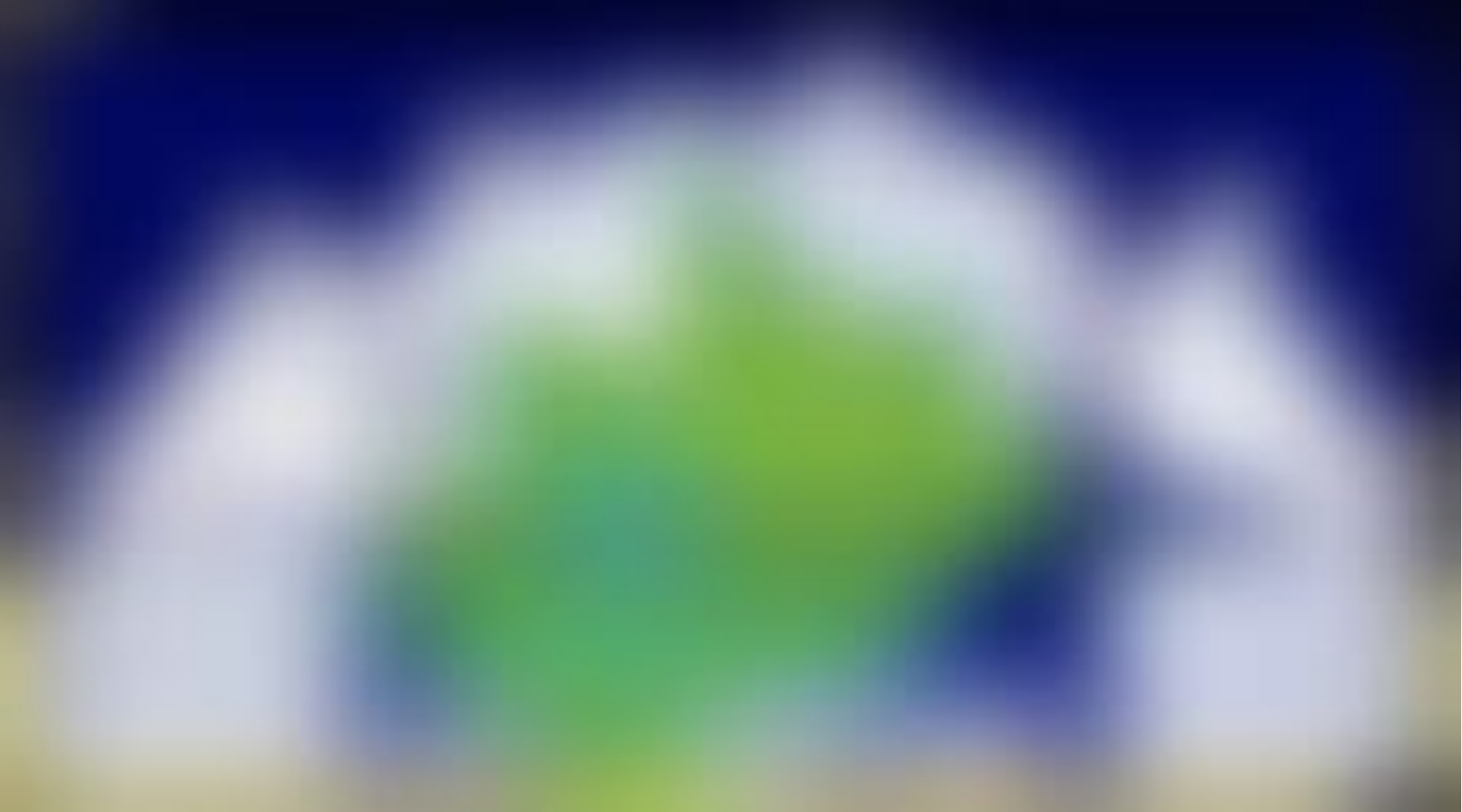} 
    \end{subfigure}
    \caption{Human-expert placements are shown on the left and results from our approach are shown on the right. The white area represents macros and the green area represents standard cells. The figures are intentionally blurred as the designs are proprietary.}
    \label{fig:visualization-plc}
\end{figure*}

\begin{table*}[ht]
\centering
\caption{Experiments to evaluate sample efficiency of Deep RL compared to Simulated Annealing (SA). We replaced our RL policy network with SA and ran 128 different SA experiments for each block, sweeping different hyper-parameters, including min and max temperature, seed, and max step size. The results from the run with minimum cost is reported. The results show proxy wirelength and congestion values for each block. Note that because these proxy metrics are relative, comparisons are only valid for different placements of the same block.}
\begin{tabular}{|c|c|c|c|c|}
\hline
        & \multicolumn{2}{c|}{Replacing Deep RL with SA in our framework} & \multicolumn{2}{c|}{Ours} \\ \hline
        & Wirelength              & Congestion              & Wirelength  & Congestion  \\ \hline

Block 1 & 0.048 & 1.21 & 0.047 & 0.87 \\ \hline
Block 2 & 0.045 & 1.11 & 0.041 & 0.93 \\ \hline
Block 3 & 0.044 & 1.14 & 0.034 & 0.96 \\ \hline
Block 4 & 0.030 & 0.87 & 0.024 & 0.78 \\ \hline
Block 5 & 0.045 & 1.29 & 0.038 & 0.88 \\ \hline
\end{tabular}
\label{table:rl-vs-sa}
\end{table*}
\subsection{Learning from Larger Datasets}

As we train on more chip blocks, we are able to speed up the training process and generate higher quality results faster. Figure \ref{fig:trainingsize} (left) shows the impact of a larger training set on performance. The training dataset is created from internal TPU blocks. The training data consists of a variety of blocks including memory subsystems, compute units, and control logic. As we increase the training set from 2 blocks to 5 blocks and finally to 20 blocks, the policy network generates better placements both at zero-shot and after being finetuned for the same number of hours. Figure \ref{fig:trainingsize} (right) shows the placement cost on the test data, as the policy network is being (pre-)trained. We can see that for the small training dataset, the policy network quickly overfits to the training data and performance on the test data degrades, whereas it takes longer for the policy network to overfit on largest dataset and the policy network pre-trained on this larger dataset yields better results on the test data. This plot suggests that as we expose the policy network to a greater variety of distinct blocks, while it might take longer for the policy network to pre-train, the policy network becomes less prone to overfitting and better at finding optimized placements for new unseen blocks.

\subsection{Visualization Insights} Figure \ref{fig:zeroshot-vs-finetuned} show the placement results for the Ariane RISC-V CPU. On the left, placements from the zero-shot policy network and on the right, placements from the finetuned policy network are shown. The zero-shot placements are generated at inference time on a previously unseen chip. The zero-shot policy network places the standard cells in the center of the canvas surrounded by macros, which is already quite close to the optimal arrangement. After finetuning, the placements of macros become more regularized and the standard cell area in the center becomes less congested.

Figure \ref{fig:visualization-plc} shows the visualized placements: on the left, results from a manual placement, and on the right, results from our approach are shown. The white area shows the macro placements and the green area shows the standard cell placements. Our method creates donut-shaped placements of macros, surrounding standard cells, which results in a reduction in the total wirelength. 

\subsection{Comparing with Baseline Methods}
 In this section, we compare our method with 3 baselines methods: Simulated Annealing, RePlAce, and human expert baselines. For our method, we use a policy pre-trained on the largest dataset (of 20 TPU blocks) and then finetune it on 5 target unseen blocks denoted by Blocks 1 to 5. Our dataset consists a variety of blocks including memory subsystems, compute units, and control logic. Due to confidentiality, we cannot disclose the details of these blocks, but to give an idea of the scale, each block contains up to a few hundred macros and millions of standard cells.

\begin{table*}[t]
\centering
\caption{Comparing our method with the state-of-the-art (RePlAce \cite{RePlAce19}) method and manual expert placements using an industry standard electronic design automation (EDA) tool. For all metrics in this table, lower is better. For placements which violate constraints on timing (WNS significantly greater than 100 ps) or congestion (horizontal or vertical congestion greater than 1\%), we render their metrics in gray to indicate that these placements are infeasible.}
\label{table:replace-comparison}
\begin{tabular}{|c|c|c|c|c|c|c|c|c|}
\hline
Name & Method & \multicolumn{2}{c|}{Timing} & Area & Power & Wirelength & \multicolumn{2}{l|}{Congestion}\\ \hline
  & & WNS (ps) & TNS (ns) & Total ($\mu m^2$) & Total (W) & (m) & H (\%) & V (\%) \\ \Xhline{4\arrayrulewidth}
Block 1 & RePlAce &  \textcolor{gray}{374} & \textcolor{gray}{233.7} & \textcolor{gray}{1693139} & \textcolor{gray}{3.70} & \textcolor{gray}{52.14} & \textcolor{gray}{1.82} & \textcolor{gray}{0.06} \\ \hline
& Manual &	136 &	47.6 & {1680790} & 3.74 & {51.12} &	{0.13} & 0.03  \\ \hline
& Ours & {84} &	{23.3} & {1681767} & {3.59} & {51.29} & 0.34 & {0.03}
\\ \Xhline{4\arrayrulewidth}
Block 2 & RePlAce  & \textcolor{gray}{97} & \textcolor{gray}{6.6} & \textcolor{gray}{785655} & \textcolor{gray}{3.52} &
\textcolor{gray}{61.07} & \textcolor{gray}{1.58} & \textcolor{gray}{0.06}\\ \hline
&Manual  &	75 & 98.1 & 830470 & 3.56 & 62.92 &	{0.23} & 0.04\\ \hline
&Ours  	 &	{59} &	170 & {694757} & {3.13} & {59.11} & 0.45 & {0.03}\\ \Xhline{4\arrayrulewidth}
 Block 3 & RePlAce  &  \textcolor{gray}{193} & \textcolor{gray}{3.9}& \textcolor{gray}{867390} & \textcolor{gray}{1.36} &
 \textcolor{gray}{18.84} &
 \textcolor{gray}{0.19} & \textcolor{gray}{0.05}\\ \hline
& Manual  &	18 &	{0.2} & 869779 & 1.42 & 20.74 &	0.22 &	0.07\\ \hline
& Ours &	{11} &	2.2 &  
{868101} &{1.38} & 20.80  & {0.04} & {0.04}  \\ 
\Xhline{4\arrayrulewidth}
Block 4 & RePlAce  & 58 & 11.2 & 944211 & 2.21 & {27.37} & 0.03 & 0.03 \\ \hline
& Manual &	58 & 17.9 & 947766 & {2.17} & 29.16 & {0.00} &{0.01} \\ \hline
& Ours & {52} & {0.7} & {942867} & {2.21} & 28.50 & 0.03 &	0.02  \\ 
\Xhline{4\arrayrulewidth}
 Block 5 & RePlAce &  156 & 254.6 & 1477283 & 3.24 &
 31.83 & 0.04 & 0.03 \\ \hline
& Manual &	107 &	{97.2} & 1480881 & {3.23} & 37.99  &	{0.00} &	{0.01}\\ \hline
& Ours   &	{68} &	141.0 & {1472302} & {3.28} & 36.59 & {0.01} &	0.03	\\ \Xhline{4\arrayrulewidth}
\end{tabular}
\end{table*}

\textbf{Comparisons with Simulated Annealing:} Simulated Annealing (SA), is known to be a powerful, but slow, optimization method. However, like RL, simulated annealing is capable of optimizing arbitrary non-differentiable cost functions. To show the relative sample efficiency of RL, we ran experiments in which we replaced it with a simulated annealing based optimizer. In these experiments, we use the same inputs and cost function as before, but in each episode, the simulated annealing optimizer places all macros, followed by an FD step to place the standard cell clusters. Each macro placement is accepted according to the SA update rule using an exponential decay annealing schedule \cite{SimulatedAnnealing}. SA takes 18 hours to converge, whereas our method takes no more than 6 hours. To make comparisons fair, we ran multiple SA experiments that sweep different hyper-parameters, including min and max temperature, seed, and max SA episodes, such that SA and RL spend the same amount of CPU-hours in simulation and search a similar number of states. The results from the experiment with minimum cost are reported in Table \ref{table:rl-vs-sa}. As shown in the table, even with additional time, SA struggles to produce high-quality placements compared to our approach, and produces placements with $14.4\%$ higher wirelength and $24.1\%$ higher congestion on average.

\textbf{Comparisons with RePlAce \cite{RePlAce19} and manual baselines:} Table \ref{table:replace-comparison} compares our results with the state-of-the-art method RePlAce \cite{RePlAce19} and manual baselines. The manual baseline is generated by a production chip design team, and involved many iterations of placement optimization, guided by feedback from a commercial EDA tool over a period of several weeks.

With respect to RePlAce, we share the same optimization goals, namely to optimize global placement in chip design, but we use different objective functions. Thus, rather than comparing results from different cost functions, we treat the output of a commercial EDA tool as ground truth. To perform this comparison, we fix the macro placements generated by our method and by RePlAce and allow a commercial EDA tool to further optimize the standard cell placements, using the tool's default settings. We then report total wirelength, timing (worst (WNS) and total (TNS) negative slack), area, and power metrics. As shown in Table \ref{table:replace-comparison}, our method outperforms RePLAce in generating placements that meet the design requirements. Given constraints imposed by the underlying semiconductor technology, placements of these blocks will not be able to meet timing constraints in the later stage of the design flow if the WNS is significantly above 100 ps or if the horizontal or vertical congestion is over 1\%, rendering some RePlAce placements (Blocks 1, 2, 3) unusable. These results demonstrate that our congestion-aware approach is effective in generating high-quality placements that meet design criteria.



RePlAce is faster than our method as it converges in 1 to 3.5 hours, whereas our results were achieved in 3 to 6 hours. However, some of the fundamental advantages of our approach are 1) our method can readily optimize for various non-differentiable cost functions, without the need to formulate closed form or differentiable equivalents of those cost functions. For example, while it is straightforward to model wirelength as a convex function, this is not true for routing congestion or timing. 2) our method has the ability to improve over time as the policy is exposed to more chip blocks, and 3) our method is able to adhere to various design constraints, such as blockages of differing shapes.

Table \ref{table:replace-comparison} also shows the results generated by human expert chip designers. Both our method and human experts consistently generate viable placements, meaning that they meet the timing and congestion design criteria. We also outperform or match manual placements in WNS, area, power, and wirelength. Furthermore, our end-to-end learning-based approach takes less than 6 hours, whereas the manual baseline involves a slow iterative optimization process with experts in the loop and can take multiple weeks. 

\subsection{Discussions}

\textbf{Opportunities for further optimization of our approach:} There are multiple opportunities to further improve the quality of our method. For example, the process of standard cell partitioning, row and column selection, as well as selecting the order in which the macros are placed all can be further optimized. In addition, we would also benefit from a more optimized approach to standard cell placement. Currently, we use a force-directed method to place standard cells due to its fast runtime. However, we believe that more advanced techniques for standard cell placement such as RePlAce \cite{RePlAce19} and DREAMPlace \cite{Dreamplace19} can yield more accurate standard cell placements to guide the policy network training. This is helpful because if the policy network has a clearer signal on how its macro placements affect standard cell placement and final metrics, it can learn to make more optimal macro placement decisions.

\textbf{Implications for a broader class of problems:} This work is just one example of domain-adaptive policies for optimization and can be extended to other stages of the chip design process, such as architecture and logic design, synthesis, and design verification, with the goal of training ML models that improve as they encounter more instances of the problem. A learning based method also enables further design space exploration and co-optimization within the cascade of tasks that compose the chip design process. 

\section{Conclusion}
In this work, we target the complex and impactful problem of chip placement. We propose an RL-based approach that enables transfer learning, meaning that the RL agent becomes faster and better at chip placement as it gains experience on a greater number of chip netlists. We show that our method outperforms state-of-the-art baselines and can generate placements that are superior or comparable to human experts on modern accelerators. Our method is end-to-end and generates placements in under 6 hours, whereas the strongest baselines require human experts in the loop and take several weeks.

\section{Acknowledgments}
This project was a collaboration between Google Research and the Google Chip Implementation and Infrastructure (CI2) Team. We would like to thank Cliff Young, Ed Chi, Chip Stratakos, Sudip Roy, Amir Yazdanbakhsh, Nathan Myung-Chul Kim, Sachin Agarwal, Bin Li, Martin Abadi, Amir Salek, Samy Bengio, and David Patterson for their help and support. 

\bibliography{example_paper}

\begin{thebibliography}{44}
\providecommand{\natexlab}[1]{#1}
\providecommand{\url}[1]{\texttt{#1}}
\expandafter\ifx\csname urlstyle\endcsname\relax
  \providecommand{\doi}[1]{doi: #1}\else
  \providecommand{\doi}{doi: \begingroup \urlstyle{rm}\Url}\fi

\bibitem[Addanki et~al.(2019)Addanki, Venkatakrishnan, Gupta, Mao, and
  Alizadeh]{Placeto18}
Addanki, R., Venkatakrishnan, S.~B., Gupta, S., Mao, H., and Alizadeh, M.
\newblock Placeto: Learning generalizable device placement algorithms for
  distributed machine learning.
\newblock \emph{CoRR}, abs/1906.08879, 2019.
\newblock URL \url{http://arxiv.org/abs/1906.08879}.

\bibitem[Agnihotri et~al.(2005)Agnihotri, Ono, and Madden]{fengshui2005}
Agnihotri, A., Ono, S., and Madden, P.
\newblock Recursive bisection placement: Feng shui 5.0 implementation details.
\newblock In \emph{Proceedings of the International Symposium on Physical
  Design}, pp.\  230--232, 01 2005.
\newblock \doi{10.1145/1055137.1055186}.

\bibitem[{Bo Hu} \& {Marek-Sadowska}(2005){Bo Hu} and
  {Marek-Sadowska}]{mfar2005}
{Bo Hu} and {Marek-Sadowska}, M.
\newblock Multilevel fixed-point-addition-based vlsi placement.
\newblock \emph{IEEE Transactions on Computer-Aided Design of Integrated
  Circuits and Systems}, 24\penalty0 (8):\penalty0 1188--1203, Aug 2005.
\newblock ISSN 1937-4151.
\newblock \doi{10.1109/TCAD.2005.850802}.

\bibitem[Brenner et~al.(2008)Brenner, Struzyna, and Vygen]{bonnplace2008}
Brenner, U., Struzyna, M., and Vygen, J.
\newblock Bonnplace: Placement of leading-edge chips by advanced combinatorial
  algorithms.
\newblock \emph{Trans. Comp.-Aided Des. Integ. Cir. Sys.}, 27\penalty0
  (9):\penalty0 1607–1620, September 2008.
\newblock ISSN 0278-0070.
\newblock \doi{10.1109/TCAD.2008.927674}.
\newblock URL \url{https://doi.org/10.1109/TCAD.2008.927674}.

\bibitem[Breuer(1977)]{MinCutBreuer1977}
Breuer, M.~A.
\newblock A class of min-cut placement algorithms.
\newblock In \emph{Proceedings of the 14th Design Automation Conference}, DAC
  ’77, pp.\  284–290. IEEE Press, 1977.

\bibitem[{Chen} et~al.(2008){Chen}, {Jiang}, {Hsu}, {Chen}, and
  {Chang}]{ntuplace32008}
{Chen}, T., {Jiang}, Z., {Hsu}, T., {Chen}, H., and {Chang}, Y.
\newblock Ntuplace3: An analytical placer for large-scale mixed-size designs
  with preplaced blocks and density constraints.
\newblock \emph{IEEE Transactions on Computer-Aided Design of Integrated
  Circuits and Systems}, 27\penalty0 (7):\penalty0 1228--1240, July 2008.
\newblock ISSN 1937-4151.
\newblock \doi{10.1109/TCAD.2008.923063}.

\bibitem[Chen et~al.(2006)Chen, Jiang, Hsu, Chen, and Chang]{NTUPlacer06}
Chen, T.-C., Jiang, Z.-W., Hsu, T.-C., Chen, H.-C., and Chang, Y.-W.
\newblock A high-quality mixed-size analytical placer considering preplaced
  blocks and density constraints.
\newblock In \emph{Proceedings of the 2006 IEEE/ACM International Conference on
  Computer-Aided Design}, ICCAD ’06, pp.\  187–192, New York, NY, USA,
  2006. Association for Computing Machinery.
\newblock ISBN 1595933891.

\bibitem[{Cheng} et~al.(2019){Cheng}, {Kahng}, {Kang}, and {Wang}]{RePlAce19}
{Cheng}, C., {Kahng}, A.~B., {Kang}, I., and {Wang}, L.
\newblock Replace: Advancing solution quality and routability validation in
  global placement.
\newblock \emph{IEEE Transactions on Computer-Aided Design of Integrated
  Circuits and Systems}, 38\penalty0 (9):\penalty0 1717--1730, 2019.

\bibitem[{Chung-Kuan Cheng} \& {Kuh}(1984){Chung-Kuan Cheng} and
  {Kuh}]{ResistiveNetwork1984}
{Chung-Kuan Cheng} and {Kuh}, E.~S.
\newblock Module placement based on resistive network optimization.
\newblock \emph{IEEE Transactions on Computer-Aided Design of Integrated
  Circuits and Systems}, 3\penalty0 (3):\penalty0 218--225, July 1984.
\newblock ISSN 1937-4151.
\newblock \doi{10.1109/TCAD.1984.1270078}.

\bibitem[{Fiduccia} \& {Mattheyses}(1982){Fiduccia} and
  {Mattheyses}]{fiduccia1982}
{Fiduccia}, C.~M. and {Mattheyses}, R.~M.
\newblock A linear-time heuristic for improving network partitions.
\newblock In \emph{19th Design Automation Conference}, pp.\  175--181, June
  1982.
\newblock \doi{10.1109/DAC.1982.1585498}.

\bibitem[Gilbert \& Pollak(1968)Gilbert and Pollak]{gilbert1968steiner}
Gilbert, E.~N. and Pollak, H.~O.
\newblock Steiner minimal trees.
\newblock \emph{SIAM Journal on Applied Mathematics}, 16\penalty0 (1):\penalty0
  1--29, 1968.

\bibitem[Hanan \& Kurtzberg(1972)Hanan and Kurtzberg]{forcedirected1972}
Hanan, M. and Kurtzberg, J.
\newblock Placement techniques.
\newblock In \emph{Design Automation of Digital Systems}, 1972.

\bibitem[{Hsu} et~al.(2011){Hsu}, {Chang}, and
  {Balabanov}]{weightedaverage2011}
{Hsu}, M., {Chang}, Y., and {Balabanov}, V.
\newblock Tsv-aware analytical placement for 3d ic designs.
\newblock In \emph{2011 48th ACM/EDAC/IEEE Design Automation Conference (DAC)},
  pp.\  664--669, June 2011.

\bibitem[Huang et~al.(2019)Huang, Xie, Fang, Yu, Ren, Fang, Chen, and
  Hu]{cnnplacer19}
Huang, Y., Xie, Z., Fang, G., Yu, T., Ren, H., Fang, S., Chen, Y., and Hu, J.
\newblock Routability-driven macro placement with embedded cnn-based prediction
  model.
\newblock In Teich, J. and Fummi, F. (eds.), \emph{Design, Automation {\&} Test
  in Europe Conference {\&} Exhibition, {DATE} 2019, Florence, Italy, March
  25-29, 2019}, pp.\  180--185. {IEEE}, 2019.

\bibitem[{Kahng} et~al.(2005){Kahng}, {Reda}, and {Qinke Wang}]{aplace22005}
{Kahng}, A.~B., {Reda}, S., and {Qinke Wang}.
\newblock Architecture and details of a high quality, large-scale analytical
  placer.
\newblock In \emph{ICCAD-2005. IEEE/ACM International Conference on
  Computer-Aided Design, 2005.}, pp.\  891--898, Nov 2005.
\newblock \doi{10.1109/ICCAD.2005.1560188}.

\bibitem[Karypis \& Kumar(1998)Karypis and Kumar]{hmetis1998}
Karypis, G. and Kumar, V.
\newblock A hypergraph partitioning package.
\newblock In \emph{HMETIS}, 1998.

\bibitem[Kernighan(1985)]{TerminalPropagation1985}
Kernighan, D.~.
\newblock A procedure for placement of standard-cell vlsi circuits.
\newblock In \emph{IEEE TCAD}, 1985.

\bibitem[{Kim} \& {Markov}(2012){Kim} and {Markov}]{complx2012}
{Kim}, M. and {Markov}, I.~L.
\newblock Complx: A competitive primal-dual lagrange optimization for global
  placement.
\newblock In \emph{DAC Design Automation Conference 2012}, pp.\  747--755, June
  2012.

\bibitem[Kim et~al.(2010)Kim, Lee, and Markov]{simpl2010}
Kim, M.-C., Lee, D.-J., and Markov, I.~L.
\newblock Simpl: An effective placement algorithm.
\newblock In \emph{Proceedings of the International Conference on
  Computer-Aided Design}, ICCAD ’10, pp.\  649–656. IEEE Press, 2010.
\newblock ISBN 9781424481927.

\bibitem[Kim et~al.(2012{\natexlab{a}})Kim, Viswanathan, Alpert, Markov, and
  Ramji]{MAPLE12}
Kim, M.-C., Viswanathan, N., Alpert, C.~J., Markov, I.~L., and Ramji, S.
\newblock Maple: Multilevel adaptive placement for mixed-size designs.
\newblock In \emph{Proceedings of the 2012 ACM International Symposium on
  International Symposium on Physical Design}, ISPD, pp.\  193–200, New York,
  NY, USA, 2012{\natexlab{a}}. Association for Computing Machinery.

\bibitem[Kim et~al.(2012{\natexlab{b}})Kim, Viswanathan, Alpert, Markov, and
  Ramji]{maple2012}
Kim, M.-C., Viswanathan, N., Alpert, C.~J., Markov, I.~L., and Ramji, S.
\newblock Maple: Multilevel adaptive placement for mixed-size designs.
\newblock In \emph{Proceedings of the 2012 ACM International Symposium on
  International Symposium on Physical Design}, ISPD ’12, pp.\  193–200, New
  York, NY, USA, 2012{\natexlab{b}}. Association for Computing Machinery.
\newblock ISBN 9781450311670.
\newblock \doi{10.1145/2160916.2160958}.
\newblock URL \url{https://doi.org/10.1145/2160916.2160958}.

\bibitem[Kirkpatrick et~al.(1983)Kirkpatrick, Gelatt, and
  Vecchi]{SimulatedAnnealing}
Kirkpatrick, S., Gelatt, C.~D., and Vecchi, M.~P.
\newblock Optimization by simulated annealing.
\newblock \emph{Science}, 220\penalty0 (4598):\penalty0 671--680, 1983.
\newblock ISSN 0036-8075.
\newblock \doi{10.1126/science.220.4598.671}.
\newblock URL \url{https://science.sciencemag.org/content/220/4598/671}.

\bibitem[Lin et~al.(2013)Lin, Chu, Shinnerl, Bustany, and Nedelchev]{polar2013}
Lin, T., Chu, C., Shinnerl, J.~R., Bustany, I., and Nedelchev, I.
\newblock Polar: Placement based on novel rough legalization and refinement.
\newblock In \emph{Proceedings of the International Conference on
  Computer-Aided Design}, ICCAD ’13, pp.\  357–362. IEEE Press, 2013.
\newblock ISBN 9781479910694.

\bibitem[Lin et~al.(2019)Lin, Dhar, Li, Ren, Khailany, and Pan]{Dreamplace19}
Lin, Y., Dhar, S., Li, W., Ren, H., Khailany, B., and Pan, D.~Z.
\newblock Dreamplace: Deep learning toolkit-enabled gpu acceleration for modern
  vlsi placement.
\newblock In \emph{Proceedings of the 56th Annual Design Automation Conference
  2019}, DAC ’19, 2019.

\bibitem[Lu et~al.(2015)Lu, Chen, Chang, Sha, Huang, Teng, and Cheng]{EPlace15}
Lu, J., Chen, P., Chang, C.-C., Sha, L., Huang, D. J.-H., Teng, C.-C., and
  Cheng, C.-K.
\newblock Eplace: Electrostatics-based placement using fast fourier transform
  and nesterov’s method.
\newblock \emph{ACM Trans. Des. Autom. Electron. Syst.}, 20\penalty0 (2), 2015.
\newblock ISSN 1084-4309.

\bibitem[{Lu} et~al.(2015){Lu}, {Zhuang}, {Chen}, {Chang}, {Chang}, {Wong},
  {Sha}, {Huang}, {Luo}, {Teng}, and {Cheng}]{EPlacemixed15}
{Lu}, J., {Zhuang}, H., {Chen}, P., {Chang}, H., {Chang}, C., {Wong}, Y.,
  {Sha}, L., {Huang}, D., {Luo}, Y., {Teng}, C., and {Cheng}, C.
\newblock eplace-ms: Electrostatics-based placement for mixed-size circuits.
\newblock \emph{IEEE Transactions on Computer-Aided Design of Integrated
  Circuits and Systems}, 34\penalty0 (5):\penalty0 685--698, 2015.

\bibitem[Lu et~al.(2016)Lu, Zhuang, Kang, Chen, and Cheng]{EPLace3D16}
Lu, J., Zhuang, H., Kang, I., Chen, P., and Cheng, C.-K.
\newblock Eplace-3d: Electrostatics based placement for 3d-ics.
\newblock In \emph{Proceedings of the 2016 on International Symposium on
  Physical Design}, ISPD ’16, New York, NY, USA, 2016. Association for
  Computing Machinery.
\newblock ISBN 9781450340397.
\newblock \doi{10.1145/2872334.2872361}.
\newblock URL \url{https://doi.org/10.1145/2872334.2872361}.

\bibitem[Nazi et~al.(2019)Nazi, Hang, Goldie, Ravi, and
  Mirhoseini]{nazi2019gap}
Nazi, A., Hang, W., Goldie, A., Ravi, S., and Mirhoseini, A.
\newblock Gap: Generalizable approximate graph partitioning framework, 2019.

\bibitem[Obermeier et~al.(2005)Obermeier, Ranke, and Johannes]{kraftwerk2005}
Obermeier, B., Ranke, H., and Johannes, F.
\newblock Kraftwerk: a versatile placement approach.
\newblock In \emph{ISPD}, pp.\  242--244, 01 2005.
\newblock \doi{10.1145/1055137.1055190}.

\bibitem[Paliwal et~al.(2019)Paliwal, Gimeno, Nair, Li, Lubin, Kohli, and
  Vinyals]{REGAL19}
Paliwal, A.~S., Gimeno, F., Nair, V., Li, Y., Lubin, M., Kohli, P., and
  Vinyals, O.
\newblock Regal: Transfer learning for fast optimization of computation graphs.
\newblock \emph{ArXiv}, abs/1905.02494, 2019.

\bibitem[{Ren-Song Tsay} et~al.(1988){Ren-Song Tsay}, {Kuh}, and {Chi-Ping
  Hsu}]{proud1988}
{Ren-Song Tsay}, {Kuh}, E.~S., and {Chi-Ping Hsu}.
\newblock Proud: a sea-of-gates placement algorithm.
\newblock \emph{IEEE Design Test of Computers}, 5\penalty0 (6):\penalty0
  44--56, Dec 1988.
\newblock ISSN 1558-1918.
\newblock \doi{10.1109/54.9271}.

\bibitem[Roy et~al.(2007)Roy, Papa, and Markov]{capo2007}
Roy, J.~A., Papa, D.~A., and Markov, I.~L.
\newblock \emph{Capo: Congestion-Driven Placement for Standard-cell and RTL
  Netlists with Incremental Capability}, pp.\  97--133.
\newblock Springer US, Boston, MA, 2007.

\bibitem[Sarrafzadeh et~al.(2003)Sarrafzadeh, Wang, and Yang]{dragon}
Sarrafzadeh, M., Wang, M., and Yang, X.
\newblock \emph{Dragon: A Placement Framework}, pp.\  57--89.
\newblock Springer, 01 2003.
\newblock ISBN 978-1-4419-5309-4.
\newblock \doi{10.1007/978-1-4757-3781-3_3}.

\bibitem[Schulman et~al.(2017)Schulman, Wolski, Dhariwal, Radford, and
  Klimov]{ppo17}
Schulman, J., Wolski, F., Dhariwal, P., Radford, A., and Klimov, O.
\newblock Proximal policy optimization algorithms, 2017.

\bibitem[Sechen \& Sangiovanni-Vincentelli(1986)Sechen and
  Sangiovanni-Vincentelli]{DAC-1986-SechenS}
Sechen, C. and Sangiovanni-Vincentelli, A.~L.
\newblock {TimberWolf3.2: a new standard cell placement and global routing
  package}.
\newblock In \emph{{DAC}}, pp.\  432--439. {IEEE Computer Society Press}, 1986.
\newblock \doi{10.1145/318013.318083}.

\bibitem[Shahookar \& Mazumder(1991)Shahookar and Mazumder]{hpwl1991}
Shahookar, K. and Mazumder, P.
\newblock Vlsi cell placement techniques.
\newblock \emph{ACM Comput. Surv.}, 23\penalty0 (2):\penalty0 143–220, June
  1991.
\newblock ISSN 0360-0300.
\newblock \doi{10.1145/103724.103725}.
\newblock URL \url{https://doi.org/10.1145/103724.103725}.

\bibitem[{Spindler} et~al.(2008){Spindler}, {Schlichtmann}, and
  {Johannes}]{kraftwerk22008}
{Spindler}, P., {Schlichtmann}, U., and {Johannes}, F.~M.
\newblock Kraftwerk2—a fast force-directed quadratic placement approach using
  an accurate net model.
\newblock \emph{IEEE Transactions on Computer-Aided Design of Integrated
  Circuits and Systems}, 27\penalty0 (8):\penalty0 1398--1411, Aug 2008.
\newblock ISSN 1937-4151.
\newblock \doi{10.1109/TCAD.2008.925783}.

\bibitem[{Tao Luo} \& {Pan}(2008){Tao Luo} and {Pan}]{dplace2008}
{Tao Luo} and {Pan}, D.~Z.
\newblock Dplace2.0: A stable and efficient analytical placement based on
  diffusion.
\newblock In \emph{2008 Asia and South Pacific Design Automation Conference},
  pp.\  346--351, March 2008.
\newblock \doi{10.1109/ASPDAC.2008.4483972}.

\bibitem[Viswanathan et~al.(2007{\natexlab{a}})Viswanathan, Nam, Alpert,
  Villarrubia, Ren, and Chu]{rql2007}
Viswanathan, N., Nam, G.-J., Alpert, C., Villarrubia, P., Ren, H., and Chu, C.
\newblock Rql: Global placement via relaxed quadratic spreading and
  linearization.
\newblock In \emph{Proceedings - Design Automation Conference}, pp.\  453--458,
  07 2007{\natexlab{a}}.
\newblock ISBN 978-1-59593-627-1.
\newblock \doi{10.1145/1278480.1278599}.

\bibitem[Viswanathan et~al.(2007{\natexlab{b}})Viswanathan, Pan, and
  Chu]{fastplace2007}
Viswanathan, N., Pan, M., and Chu, C.
\newblock \emph{FastPlace: An Efficient Multilevel Force-Directed Placement
  Algorithm}, pp.\  193--228.
\newblock Springer, 01 2007{\natexlab{b}}.
\newblock \doi{10.1007/978-0-387-68739-1_8}.

\bibitem[William et~al.(2001)William, Ross, and Lu]{logsumexp2001}
William, N., Ross, D., and Lu, S.
\newblock Non-linear optimization system and method for wire length and delay
  optimization for an automatic electric circuit placer.
\newblock In \emph{Patent}, 2001.

\bibitem[Zhang \& Chen(2018)Zhang and Chen]{zhang2018link}
Zhang, M. and Chen, Y.
\newblock Link prediction based on graph neural networks, 2018.

\bibitem[Zhiyao Xie Duke~Univeristy(2018)]{RouteNet18}
Zhiyao Xie Duke~Univeristy, Durham, N. U. . Y.-H. H. . G.-Q. F. . H. R. . S.-Y.
  F. . Y. C. . J.~H.
\newblock Routenet: Routability prediction for mixed-size designs using
  convolutional neural network.
\newblock In \emph{IEEE/ACM International Conference on Computer-Aided Design
  (ICCAD}, 2018.

\bibitem[Zhou et~al.(2019)Zhou, Roy, Abdolrashidi, Wong, Ma, Xu, Zhong, Liu,
  Goldie, Mirhoseini, and Laudon]{zhou2019gdp}
Zhou, Y., Roy, S., Abdolrashidi, A., Wong, D., Ma, P.~C., Xu, Q., Zhong, M.,
  Liu, H., Goldie, A., Mirhoseini, A., and Laudon, J.
\newblock Gdp: Generalized device placement for dataflow graphs, 2019.

\end{thebibliography}
\bibliographystyle{icml2020}
\end{document}